\pdfoutput=1

\documentclass[11pt]{article}

\usepackage[]{acl}

\usepackage{times}
\usepackage{latexsym}

\usepackage[T1]{fontenc}
\usepackage[utf8]{inputenc}

\usepackage{microtype}

\usepackage{amsmath}
\usepackage{graphicx}
\usepackage{subcaption}
\usepackage[normalem]{ulem}
\useunder{\uline}{\ul}{}
\usepackage{threeparttable}
\usepackage{lipsum}
\usepackage{enumerate}
\usepackage{booktabs}
\usepackage{multirow}
\usepackage{amssymb}
\usepackage{bookmark}

\title{Is Reference Necessary in the Evaluation of NLG Systems? 

When and Where?}
\author{Shuqian Sheng\textsuperscript{1}, Yi Xu\textsuperscript{1}, Luoyi Fu\textsuperscript{1\thanks{* Luoyi Fu is the corresponding author.}}, Jiaxin Ding\textsuperscript{1}, \\\ \textbf{Lei Zhou\textsuperscript{1}, Xinbing Wang\textsuperscript{1}, Chenghu Zhou\textsuperscript{2}}\\
\textsuperscript{1}Shanghai Jiao Tong University, Shanghai, China\\
\textsuperscript{2}IGSNRR, Chinese Academy of Sciences, Beijing, China\\
  {\tt \{susisheng, yixu98, yiluofu\}@sjtu.edu.cn} \\  }

\begin{document}
\maketitle
\begin{abstract}
    The majority of automatic metrics for evaluating NLG systems are reference-based. However, the challenge of collecting human annotation results in a lack of reliable references in numerous application scenarios. Despite recent advancements in reference-free metrics, it has not been well understood when and where they can be used as an alternative to reference-based metrics. In this study, by employing diverse analytical approaches, we comprehensively assess the performance of both metrics across a wide range of NLG tasks, encompassing eight datasets and eight evaluation models. Based on solid experiments, the results show that reference-free metrics exhibit a higher correlation with human judgment and greater sensitivity to deficiencies in language quality. However, their effectiveness varies across tasks and is influenced by the quality of candidate texts. Therefore, it's important to assess the performance of reference-free metrics before applying them to a new task, especially when inputs are in uncommon form or when the answer space is highly variable. Our study can provide insight into the appropriate application of automatic metrics and the impact of metric choice on evaluation performance.
  
\end{abstract}

\section{Introduction}

\begin{figure}[!t]
  \centering
  \vspace{2em}
  \includegraphics[width=0.48\textwidth]{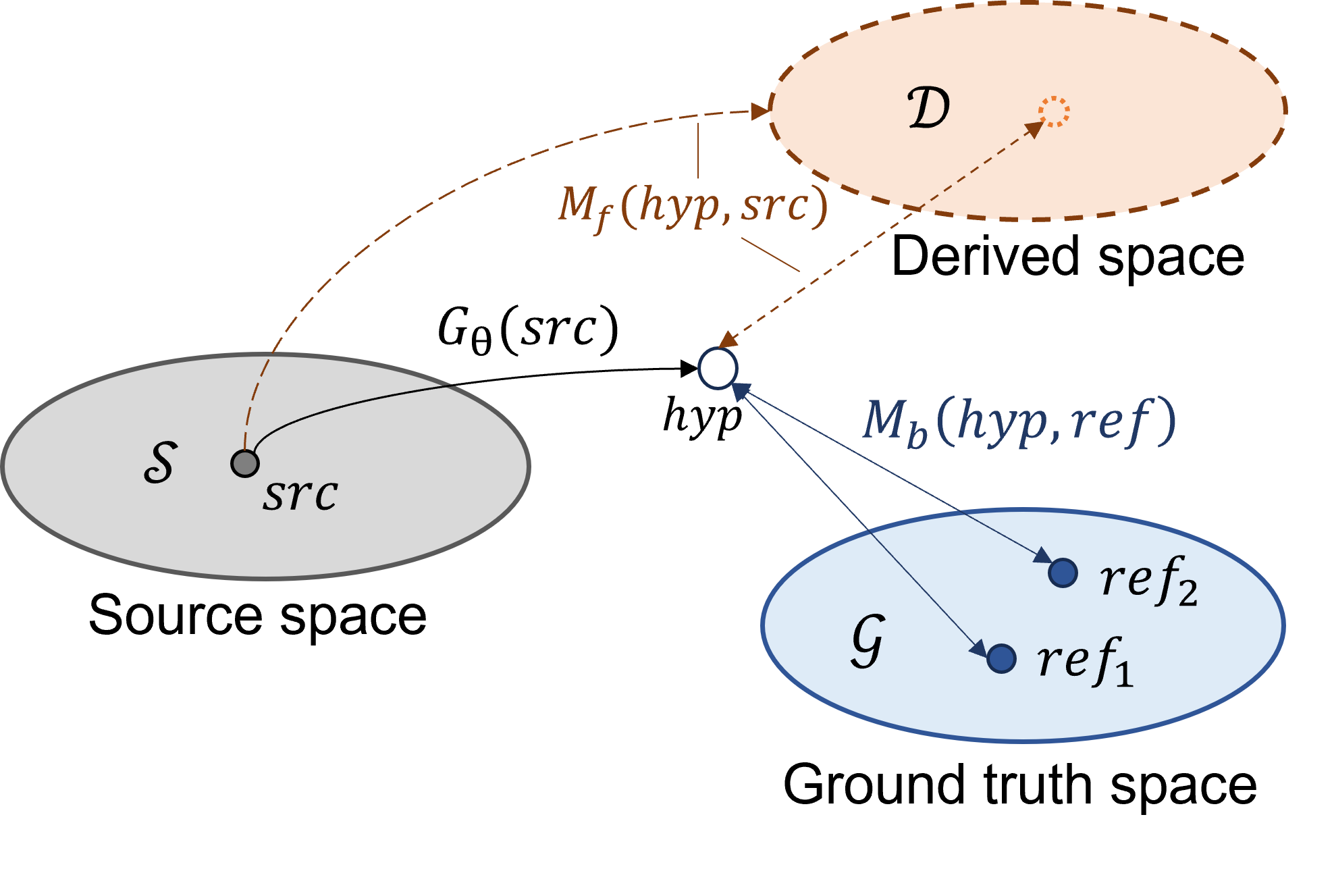}
  \caption{Evaluation mechanism of automatic evaluation metrics. Reference-based metrics measure the similarity between $hyp$ and $ref$s, while reference-free metrics instead measure how likely the $hyp$ is in the derived space $\mathcal{D}$.}
\label{fig:metric_intro}
\vspace{-0.5em}
\end{figure}

Automatic evaluation metrics for generated texts play a crucial role in the development of Natural Language Generation(NLG) techniques. Most commonly used metrics are reference-based~\cite{Papineni-2002-bleu, banerjee-lavie-2005-meteor, zhangbertscore, zhao-etal-2019-moverscore}. 
Such metrics provide evaluation results by measuring the similarity between text and human-written references~\cite{gehrmann2023repairing}, 
which are widely applied in various evaluation tasks.

However, in the era of Large Language Models (LLMs), we are witnessing the emergence of LLMs with varying parameters and domain-specific variations. 
Although their performance can be tested using standard benchmarks, due to the lack of ground truth reference texts, evaluating the generated texts of language models in specific user-oriented scenarios with reference-based metrics is challenging. 
As a result, the assessment of coherence, consistency, fluency, and other criteria of the language model's output demands substantial time and cognitive resources.

In recent years, many reference-free metrics have been proposed as a potential solution to the aforementioned challenges~\cite{NEURIPS2021_bartscore, fu2023gptscore, zhong-etal-2022-unieval}. The evaluation procedure of reference-free metrics can be viewed as a generative process, using an underlying generation model to assess other models~\cite{deutsch-etal-2022-limitations}, without any reliance on human annotations. The evaluation processes of both reference-based and reference-free metrics are illustrated in Figure~\ref{fig:metric_intro}, and we will give a formal definition in Section~\ref{sec:standard_eval}.

Although in some tasks the evaluation results of reference-free metrics have shown a higher correlation with human assessment~\cite{freitag-etal-2021-wmt-results}, \textbf{when and where} can they be used as a substitute for reference-based metrics is still not well understood. In order to figure out the answer, in this study, we mainly focus on the following questions:
\begin{itemize}
  \item On which tasks and criteria can reference-free metrics outperform reference-based metrics? 
  \item In case reference is necessary, what is the reason behind such a requirement?
  \item Considering the advantages and limitations of each metric, how can we better utilize automatic evaluation techniques?
\end{itemize}

Specifically, we employ three task-independent criteria: coherence, consistency, and fluency, to comprehensively evaluate the differences between the two types of metrics across various tasks. Three reference-free and five reference-based metrics are included, which are tested on eight datasets spanning three tasks: summarization, data-to-text, and dialogue. 

Our experiments reveal that,
(\romannumeral1) Regarding the first question, reference-free metrics exhibit a stronger correlation with human judgment on all three criteria and almost all tasks, and they are more sensitive to fluency and coherence deficiencies.
(\romannumeral2) As for the second question, the performance of reference-free metrics is constrained by underlying models. The evaluation effectiveness of reference-free metrics could vary across tasks and is influenced by the quality of candidate texts.
(\romannumeral3) Addressing the final question, it is crucial to assess the performance of reference-free metrics before applying them to a new task. They are capable of recognizing texts with poor quality, but may not be able to evaluate high-quality candidates.

Our contribution includes: 
\begin{itemize}
    \item We thoroughly investigate the performances of reference-free and reference-based metrics with numerous experiments, revealing their inherent advantages and limitations.
    \item We find out reference-free metrics have better performance but are limited by application scenarios, and provide possible explanations, regarding questions 1 and 2.
    \item We offer guidance on the judicious use of automatic metrics to ensure the integrity of evaluations and to prevent any potential misdirection, addressing question 3.
\end{itemize}

\section{Preliminary}

\subsection{Criteria}
In this study, we focus on three criteria: coherence, consistency, and fluency. Below, we present the definitions applied in this research.

\paragraph{Coherence}
Coherence, following~\citet{dang2005overview}, assesses whether models produce a well-structured and organized body of text based on the given task, avoiding a mere compilation of related information.

\paragraph{Consistency}
Consistency, in line with~\citet{honovich-etal-2022-true-evaluating}, evaluates whether all factual information in the output text aligns with the content provided in the input.

\paragraph{Fluency}
Fluency, referring to~\citet{kann2018sentence-fluency}, measures how naturally a sentence is perceived by humans. In some cases, fluency is also denoted as naturalness, grammaticality, or readability.

\subsection{Standard Evaluation}
\label{sec:standard_eval}

As illustrated in Figure~\ref{fig:metric_intro}, the conditional text generation process takes a source text $src$ as input and produces a hypothesis text $hyp$ as output, based on the generation function $G$. 
Here, $src$ represents an instance sampled from the source space $\mathcal{S}$. The goal of the evaluation metric is to impartially assess the quality of $hyp$, usually in the form of a score. 
When provided with an input text, one approach to obtaining the standard response is to collect answers from expert human annotators, which can be regarded as a sample from the ground truth space $\mathcal{G}$. We denote such human-written ground truth as a reference, represented as $ref$. Depending on whether the presence of $ref$ is required for the evaluation process, automatic metrics can be categorized into two types: reference-based and reference-free. 

Reference-based metrics measure the similarity between $hyp$ and one or multiple $ref$s, and a $hyp$ more similar to $ref$ is considered to be better~\cite{gehrmann2023repairing}. We denote the function used in similarity measurement as $M_b(\cdot)$.

\begin{equation}
\label{eq:score_ref_based}
    s_{b} = M_b(ref, hyp) .
\end{equation}

On the contrary, reference-free metrics are independent of text $ref$ but usually require $src$ as the input. Reference-free metrics can be viewed as a generation model, conducting evaluation based on an underlying inference procedure~\cite{deutsch-etal-2022-limitations}. As a sample from ground truth space is not available, for each given $src$, such metrics instead build up a derived space $\mathcal{D}$, from the knowledge stored in the underly models, and measure how likely the $hyp$ is in the derived space. Depending on different evaluation scenarios, the derived space can vary. We denote the metric function for reference-free metrics as $M_f(\cdot)$, and the output score satisfies the following equation: 

\begin{equation}
\label{eq:score_ref_free}
    s_f= M_f(src, hyp) .
\end{equation}

It is worth noting that, when evaluating a criterion indifferent to contextual information, some metrics do not require $src$ as the input. Though $src$ is optional, it does not conflict with Equation~\ref{eq:score_ref_free}. Another point is, though we describe the working flow of metrics with equations, metrics are calculations for scoring $hyp$s, instead of one-to-one mathematical mapping.

\subsection{Methods of Meta-evaluation}
In this study, we adopt the following methods to evaluate the effectiveness of automatic metrics.
\subsubsection{Correlations with Human}
\label{sec:correlation}
One of the most common methods for automatic metrics assessment is to measure the correlation between human judgment and metrics score, as human judgment is still the gold-standard approach to text evaluation~\cite{NEURIPS2021_bartscore}. Correlation functions used in this work include \textit{Spearman Correlation}~\cite{Spearman1987ThePA},\textit{ Pearson Correlation}~\cite{benesty2009pearson} and \textit{Kendall's Tau}~\cite{KENDALL1938}. All correlation scores are calculated at the sample level. To be specific, the sample-level correlation is defined as:
\begin{equation}
    \begin{split}
        correlation = \rho([s_1, s_2, \ldots, s_N], \\
        [h_1, h_2, \ldots, h_N]),
    \end{split}
\end{equation} where $\rho$ is the correlation function,  $s_i$ is the metric score of the i-th sample in a certain dataset, and $h_i$ is the corresponding human judgment.

\subsubsection{Criterion-level Analysis}
A single overall correlation score with human judgment may not comprehensively reflect the effectiveness of the metric~\cite{nimah-etal-2023-nlg}. Therefore, an analysis that specifically focuses on individual criteria is an important supplementary approach.

Intuitively, a metric capable of assessing a specific criterion should be able to distinguish sentences of different quality on that certain criterion. We adopt the perturbation detection test and Kolmogorov-Smirnov (KS) score for criterion-level analysis.

\paragraph{Perturbation Detection Test}
Perturbation detection tests help explore whether a metric can discern the quality drop of texts. We employ perturbation techniques outlined in~\citet{sai-etal-2021-perturbation} for criteria fluency and coherence, in order to measure metrics' ability to discern perturbed sentences from the original ones.

To be specific, we represent the score of perturbed text generated by metric $M$ as $\hat{s}$, and the score of the corresponding original text as $s$. As the quality of perturbed sentences is diminished by manual perturbation, ideally, for a competent metric, it's expected that $s > \hat{s}$ is true,  We employ the proportion of text pairs where $s > \hat{s}$ as a statistical measure to evaluate metric $M$'s ability to detect perturbations.

\paragraph{Kolmogorov-Smirnov Score}
Following the analysis method proposed in~\cite{nimah-etal-2023-nlg}, we utilize the Kolmogorov-Smirnov (KS) test as a statistical index to evaluate metrics' ability to distinguish sentences from different groups. The definition of KS score is as follows:

\begin{equation}
    KS_M = \sup_{s} |F_A(x) - F_B(x)|
\end{equation}

Here, $F_A$ and $F_B$ correspond to the empirical cumulative density functions of scores produced by metric $M$ for sentence groups $A$ and $B$, where groups $A$ and $B$ consist of sentences with varying qualities. $KS_M = 0$ means the distributions of $A$ and $B$ are identical, indicating that $M$ has poor performance in separating high-quality and low-quality texts.

\subsubsection{Stability Analysis}
\label{sec:system_level_analysis}
An eligible metric should be stable when applied to evaluate texts generated by different systems. To investigate if the effectiveness of metrics fluctuates when they are used on systems of varying quality, we utilize the meta-correlation index proposed by~\citet{shen2023large}. 

First, the quality of a system is measured by the average human score for all candidate sentences generated by the system, as shown in equation~\ref{eq:system_quality}:
\begin{equation}
\label{eq:system_quality}
    Q_i = \frac{1}{N}\sum_{j=1}^{N}h_{i,j},
\end{equation} where $N$ represents the number of candidate sentences generated by system $i$, and $h_{i,j}$ is the relevant human judgment.

The performance of metric $M$ on a specific system $i$ is assessed by the correlation between the metric score and human judgment of the system's output. 
\begin{equation}
    \begin{split}
        P_i = \rho([s_{i,1}, s_{i,2}, \ldots, s_{i,N}], \\
     [h_{i,1}, h_{i,2}, \ldots, h_{i,N}]),
    \end{split}
    \label{eq:system_correlation}
\end{equation} where $s_{i,j}$ is the metric score of the $j$-th sentence generated by system $i$.

Finally, the meta-correlation of metric $M$ is calculated on all $k$ system:
\begin{equation}
    M = \rho([Q_1, Q_2, \ldots, Q_k], [P_1, P_2, \ldots, P_k])
\end{equation}

\section{Experiments}
In this section, we first evaluate the performance of metrics on different datasets and criteria. Then, we conduct perturbation experiments to examine metrics' sensitivity concerning sentence defects and employ the KS score for further criterion-level analysis. Finally, we use the meta-correlation index to explore the stability of metric performance in relation to candidate quality.

\subsection{Metrics}
\subsubsection{Reference-free Metrics}
In this study, we select three popular reference-free metrics for analysis. GPTScore uses conditional probability to evaluate the quality of given texts~\cite{fu2023gptscore}. We use checkpoint "gpt2-large"~\cite{radford2019language}. BARTScore views the evaluation process as a generation problem, measuring how likely a target text can be generated based on the given inputs~\cite{NEURIPS2021_bartscore}, and we use the faithfulness-based variant of BARTScore. UniEval views the evaluation task as a Boolean Question~\cite{zhong-etal-2022-unieval}. We adopt the checkpoint "summarization" for evaluation. We also take the index "overall" for assessment on each criterion,  which is denoted as UniEval\_all. Apart from fluency evaluation with UniEval, which only requires $hyp$, the other evaluation process accepts $src$ and $hyp$ as input.

\subsubsection{Reference-based Metrics}
We select five common reference-based metrics for analysis. BLEU~\cite{Papineni-2002-bleu}, ROUGE~\cite{lin2004rouge} and METEOR\cite{banerjee-lavie-2005-meteor} provide evaluation results by calculating the statical index of n-gram overlap between $ref$ and $hyp$. For ROUGE, we use ROUGE-2. BERTScore ~\cite{zhangbertscore} and MoverScore~~\cite{zhao-etal-2019-moverscore} both produce the evaluation result by measuring the similarity of embeddings between $hyp$ and the $ref$. Specifically, we use the "deberta-xlarge-mnli" checkpoint~\cite{he2021deberta} for BERTScore. All reference-based metrics accept $ref$ and $hyp$ as inputs.
 (See more implementation details in Appendix~\ref{app:implementation}).

\subsection{Datasets}
% Please refer to the Appendix for more details.

We use eight datasets related to task summarization, data-to-text, and dialogue. 
Each dataset comprises samples containing the following components: source text $src$, reference text $ref$, system output $hyp$, and human judgments across various dimensions. All texts within these datasets are composed in English.
On the summarization task, we select datasets SummEval~\cite{fabbri-etal-2021-summeval}, Newsroom~\cite{grusky2018newsroom}, and QAGS~\cite{wang-etal-2020-asking}. On data-to-text task, we select SFHOT and SFRES~\cite{wen-etal-2015-semantically}, WebNLG~\cite{shimorina2019webnlg}, and BAGEL~\cite{mairesse-etal-2010-bagel}.
Specifically,  we utilize the resource assembled by \citet{NEURIPS2021_bartscore} for the datasets Newsroom, SummEval, QAGS, SFHOT, and SFRES, and resource collected by \citet{Scialom2021BEAMetricsAB} for dataset WebNLG.
On the dialogue task, we select USR~\cite{mehri-eskenazi-2020-usr}. The USR dataset consists of 2 parts, which are developed based on the Topical-Chat dataset and the Persona-Chat dataset separately. We note them USR-Topical and USR-Persona respectively. Features contained in datasets are listed in Table~\ref{tab:dataset_criteria}. Please refer to Appendix~\ref{app:implementation} for more details.\textbf{} 

\begin{table}[!h]
    \centering
    \small
    \begin{tabular}{lcccc}
    \toprule
                             & \textbf{COH}             & \textbf{CON}        & \textbf{FLU}       & \textbf{REF}        \\
    \midrule                        
    \multicolumn{1}{l}{\textbf{summarization}} & &            &            &            \\
    
    -Newsroom                 & \checkmark      &            & \checkmark & \checkmark \\
    -QAGS                     &                 & \checkmark &            & \checkmark \\
    -SummEval                 & \checkmark      & \checkmark & \checkmark & \checkmark \\
    \midrule
    \textbf{data-to-text}       &                 &            &            &            \\
    
    -BAGEL                    &              &  & \checkmark & \checkmark \\
    -SFHOT                    &                 & \checkmark & \checkmark & \checkmark \\
    -SFRES                    &                 & \checkmark & \checkmark & \checkmark \\
    -WebNLG                   &                 &            & \checkmark & \checkmark \\
    
    \midrule
    \multicolumn{2}{l}{\textbf{dialogue}}  &            &            &            \\
    -USR-Persona                    & \checkmark      &            & \checkmark &  \checkmark  \\
    -USR-Topical                      & \checkmark      &            & \checkmark &  \checkmark  \\
    \bottomrule
  
  \end{tabular}
  \caption{Datasets and available features.}
  \label{tab:dataset_criteria}

\end{table}

% spearman all
\begin{table*}[!h]
    \small
    \centering
    \begin{tabular}{l@{\hspace{0.12cm}}c@{}c@{\hspace{0.1cm}}c@{\hspace{0.1cm}}c@{\hspace{0.1cm}}c@{\hspace{0.1cm}}|c@{\hspace{0.1cm}}c@{\hspace{0.1cm}}c@{\hspace{0.1cm}}c@{\hspace{0.1cm}}c}
      \toprule
      &           &                \multicolumn{4}{c|}{\textbf{Reference-free}}        & \multicolumn{5}{c}{\textbf{Reference-based}}           \\ 
                               \cmidrule{3-11}
                         &             & \textbf{GPTScore} & \textbf{BARTScore} & \textbf{UniEval} & \textbf{UniEval\_all} & \textbf{MoverScore} & \textbf{BERTScore} & \textbf{ROUGE} &\textbf{Meteor} & \textbf{BLEU}   \\
    \midrule
    \multirow{4}{*}{\textbf{COH}}    & Newsroom           & 0.595    & \textbf{0.623}     & 0.458   & 0.486        & 0.091      & 0.221     & 0.081     & 0.198  & -0.201 \\
                         & SummEval           & 0.412    & 0.408     & \textbf{0.592}   & 0.538        & 0.154      & 0.333     & 0.153     & 0.134  & 0.125  \\
                         & USR\_Persona\ & 0.046    & 0.006     & 0.221   & 0.185        & 0.237      & \textbf{0.260}     & 0.097     & 0.179  & -0.041 \\
                         & USR\_Topic\   & 0.072    & 0.046     & \textbf{0.380}   & 0.296        & 0.260      & 0.309     & 0.253     & 0.276  & -0.172 \\
    \midrule
    \multirow{5}{*}{\textbf{CON}}    & QAGS\_CNN          & 0.583    & \textbf{0.680}     & 0.618   & 0.633        & 0.353      & 0.507     & 0.418     & 0.326  & 0.082  \\
                         & QAGS\_XSUM         & 0.081    & 0.159     & \textbf{0.387}   & 0.344        & 0.052      & -0.057    & 0.129     & -0.015 & -0.164 \\
                         & SFHOT              & 0.219    & 0.222     & 0.196   & \textbf{0.270}        & 0.201      & 0.221     & 0.088     & 0.069  & -0.106 \\
                         & SFRES              & 0.271    & 0.254     & 0.213   & \textbf{0.283}        & 0.172      & 0.184     & 0.108     & 0.175  & -0.073 \\
                         & SummEval           & 0.355    & 0.334     &\textbf{0.435}   & 0.429        & 0.146      & 0.200     & 0.069     & 0.152  & 0.048  \\
    \midrule
                         \multirow{8}{*}{\textbf{FLU}} & BAGEL              & 0.152    & 0.241     & \textbf{0.309}   & \textbf{0.309}        & 0.187      & 0.247     & 0.152     & 0.109  & 0.193  \\
                         & Newsroom           & 0.565    & \textbf{0.596}     & 0.443   & 0.516        & 0.046      & 0.182     & 0.051     & 0.157  & -0.163 \\
                         & SFHOT              & 0.135    & 0.164     & 0.312   & \textbf{0.324}        & 0.155      & 0.164     & 0.042     & 0.015  & -0.054 \\
                         & SFRES              & 0.229    & 0.226     & \textbf{0.332}   & 0.323        & 0.154      & 0.183     & 0.081     & 0.143  & 0.100  \\
                         & SummEval           & 0.288    & 0.285     &\textbf{0.451}   & 0.434        & 0.122      & 0.194     & 0.044     & 0.090  & -0.015 \\
                         & USR\_Persona\ & -0.030   & 0.034     & 0.239   & \textbf{0.367}        & 0.116      & 0.322     & 0.112     & 0.073  & -0.124 \\
                         & USR\_Topic\   & 0.087    & 0.027     & 0.302   & \textbf{0.395}        & 0.186      & 0.292     & 0.169     & 0.200  & -0.093 \\
                         & WebNLG             & 0.072    & 0.330     & 0.521   & \textbf{0.565}        & 0.429      & 0.499     & 0.277     & 0.332  & 0.318 \\
                         
    \bottomrule
    \end{tabular}
    \caption{Each row represents the \textbf{Spearman's correlations} of different metrics with human judgments on a dataset. Coherence, consistency, and fluency are written in abbreviations COH, CON, and FLU respectively. The \textbf{bold} scores represent the highest correlation results for each task on each criterion.}
    \label{tab:spearman_all}
    \vspace{-0.15cm}
\end{table*}

\subsection{Correlations with Human}
\label{sec:standard_correlation_test}
We follow the standard procedure to obtain the evaluation result of each metric, as depicted in Section~\ref{sec:correlation}. No fine-tuning is performed during experiments.
The Spearman correlations between scores generated by automatic metrics and human judgment are shown in Table~\ref{tab:spearman_all}. See Table~\ref{tab:pearson_all} and Table~\ref{tab:kendalltau_all} in the Appendix~\ref{app:supplementary} for corresponding results of Pearson correlation and Kendall's Tau.

The outcomes show that reference-free metrics \textbf{outperform} reference-based metrics across various datasets and evaluation criteria. UniEval and BARTScore achieve the highest scores in 16 test experiments. GPTScore also outperforms five reference-based metrics in most tasks. 
The poor performance of reference-based metrics may be attributed to their high dependence on the selection of $ref$. Thorough case study, we observe that scores yielded by $ref$s written in different sentence structures could vary greatly, even when they contain the same meaning. Therefore, the structure of datasets also influences the results. 
For example, in data-to-text tasks, datasets SFRES and SFHOT contain $hyps$ from handcrafted NLG systems, which are more formulaic and differ from $ref$s, while $ref$s in WebNLG are similar to the $hyp$s. The performance thus exhibits a great variation, with the latter having comparatively better performance.
Please refer to the Appendix~\ref{app:ref_influence} for more details.

We also observe that, the performance of some reference-free metrics on different tasks \textbf{exhibits significant variation}. That is, their advantages over ref-based metrics are not consistent across tasks. In the dialogue task, apart from UniEval, the performance of other reference-free metrics is worse than reference-based metrics. In data-to-text tasks, their advantage is not so pronounced. 

One reason could be that, without reference, the performance of reference-free metrics completely depends on how accurate the derived answer space is, which relies on the generation ability of the underlying model. When the underlying model is not able to handle the specific type of input, the performance of reference-free metrics will drop. In the case of dialogue datasets, the answers of each $src$ exhibit great diversity, and the derived space may not be able to cover all possible responses. Data-to-text tasks utilize structural input, whose meanings are more obscure than input written in natural language, which also causes difficulty for metrics to perform reliable assessments. 

In such cases, employing reference-based metrics for evaluation appears to be a better approach. Another possible solution could be developing source-free metrics, i.e. metrics that do not require $src$ as input. For example, UniEval only uses $hyp$ for fluency evaluation and maintains a relatively higher correlation score, implying that for criteria that do not rely on contextual information or task-specific scenarios, the inclusion of $src$ may be unnecessary, and such source-free metrics could remain unaffected by the input's structure. 

\label{sec:ks_test}
% heatmap
\begin{figure*}[t]
    \begin{minipage}[t]{0.33\textwidth} 
        \centering
        \includegraphics[width=\linewidth]{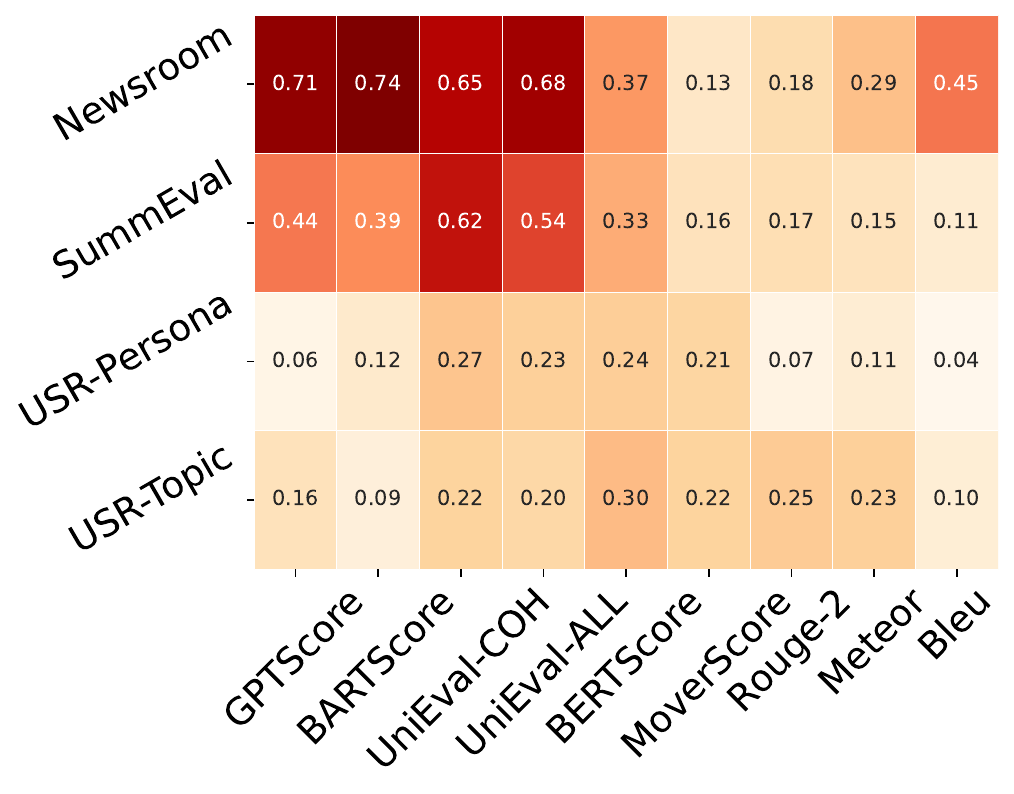}
        \subcaption{Coherence}
        \label{fig:ks-coh}
    \end{minipage}%
    \begin{minipage}[t]{0.33\textwidth}
        \centering
        \includegraphics[width=\linewidth]{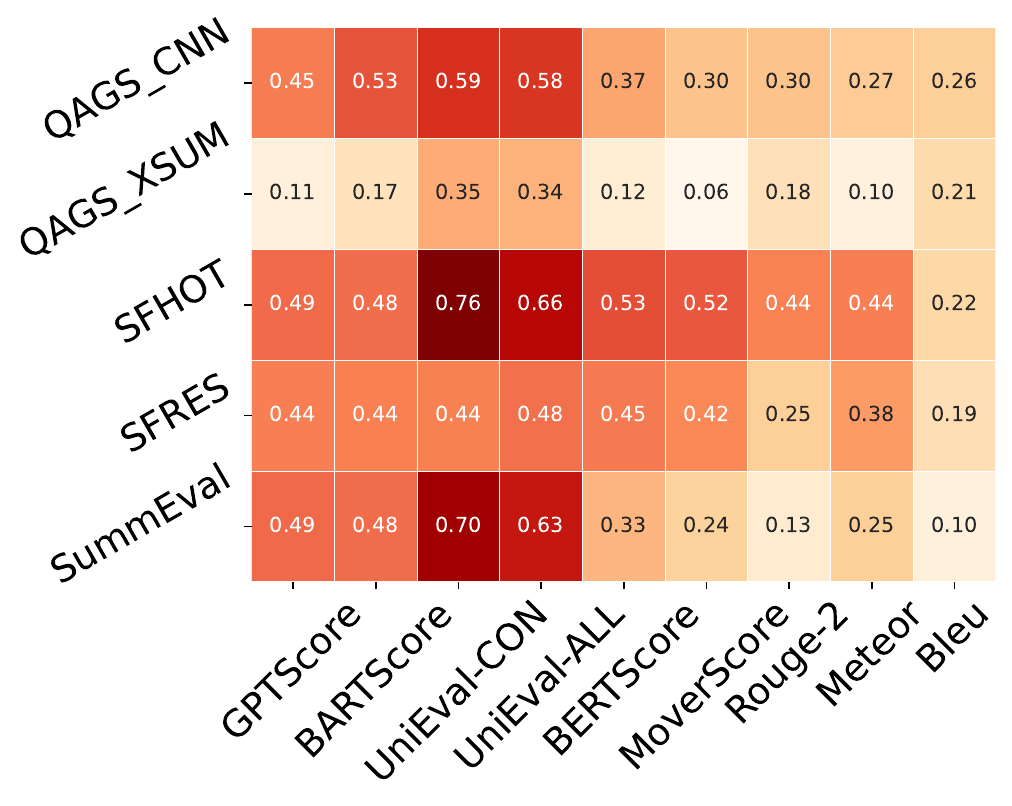}
        \subcaption{Consistency}
        \label{fig:ks-con}
    \end{minipage}
    \begin{minipage}[t]{0.33\textwidth} 
        \centering
        \includegraphics[width=\linewidth]{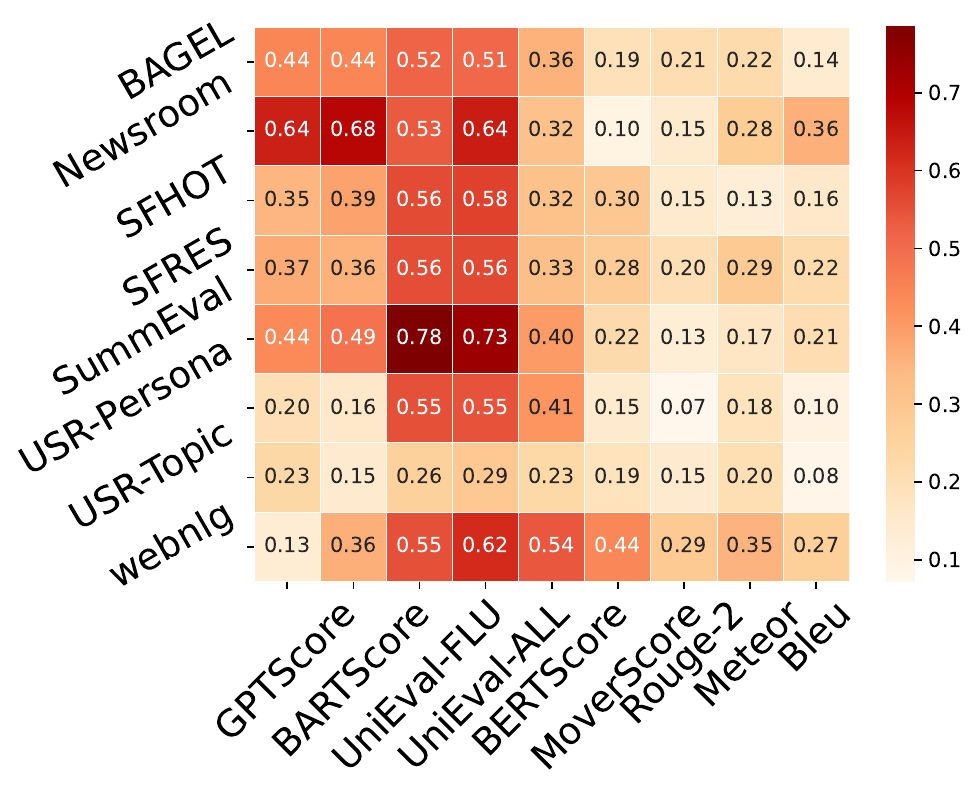}
        \subcaption{Fluency}
        \label{fig:ks-flu}
    \end{minipage}
    \caption{Heatmap of \textbf{Kolmogorov-Smirnov} (KS) score on distinguishing performance of high-quality and low-quality \textit{hyp}. The number on heatmaps represents the KS score of each metric on distinguishing high and low quality \textit{hyp} on each dataset. The range of KS score is $[0,1]$. The higher the score, the better the performance is.}
    \label{fig:ks_heatmap}
\end{figure*}

\subsection{Perturbation Experiments}
\label{sec:perturb}  

We perform perturbation experiments on the SummEval and Newsroom datasets, focusing on criteria coherence and fluency. In these experiments, we apply perturbations to the $hyp$ in each dataset and assess the resulting perturbed text with each metric, obtaining assessment scores $\hat{s}$. We exclude samples that have human evaluation judgment scores below 3 to ensure the quality of the original $hyp$. 

\begin{figure}[h]
    \begin{minipage}[h]{0.499\textwidth} 
        \centering
        \includegraphics[width=\linewidth]{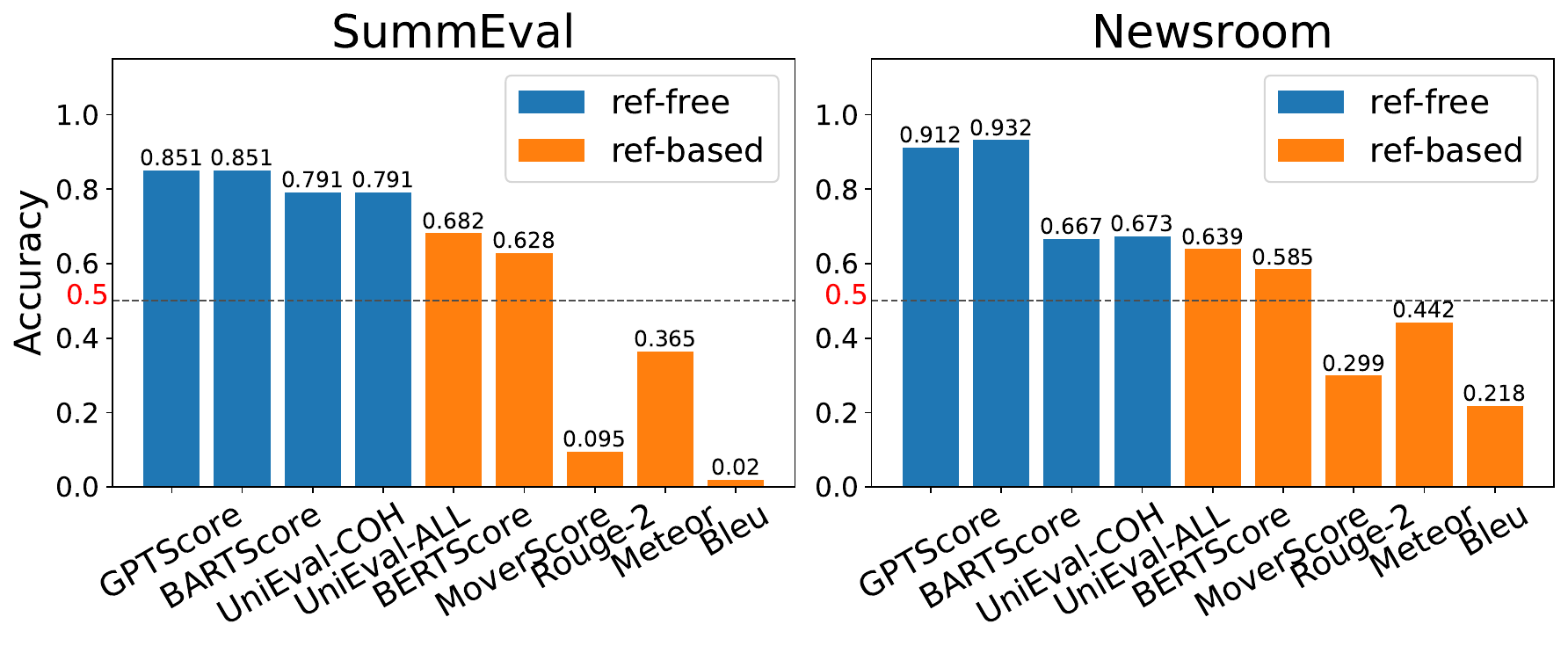}
        \subcaption{Coherence}
        \label{fig:coh_perturb_acc}

    \end{minipage}%
    \hfill

    \begin{minipage}[h]{0.499\textwidth}
        \centering
        \includegraphics[width=\linewidth]{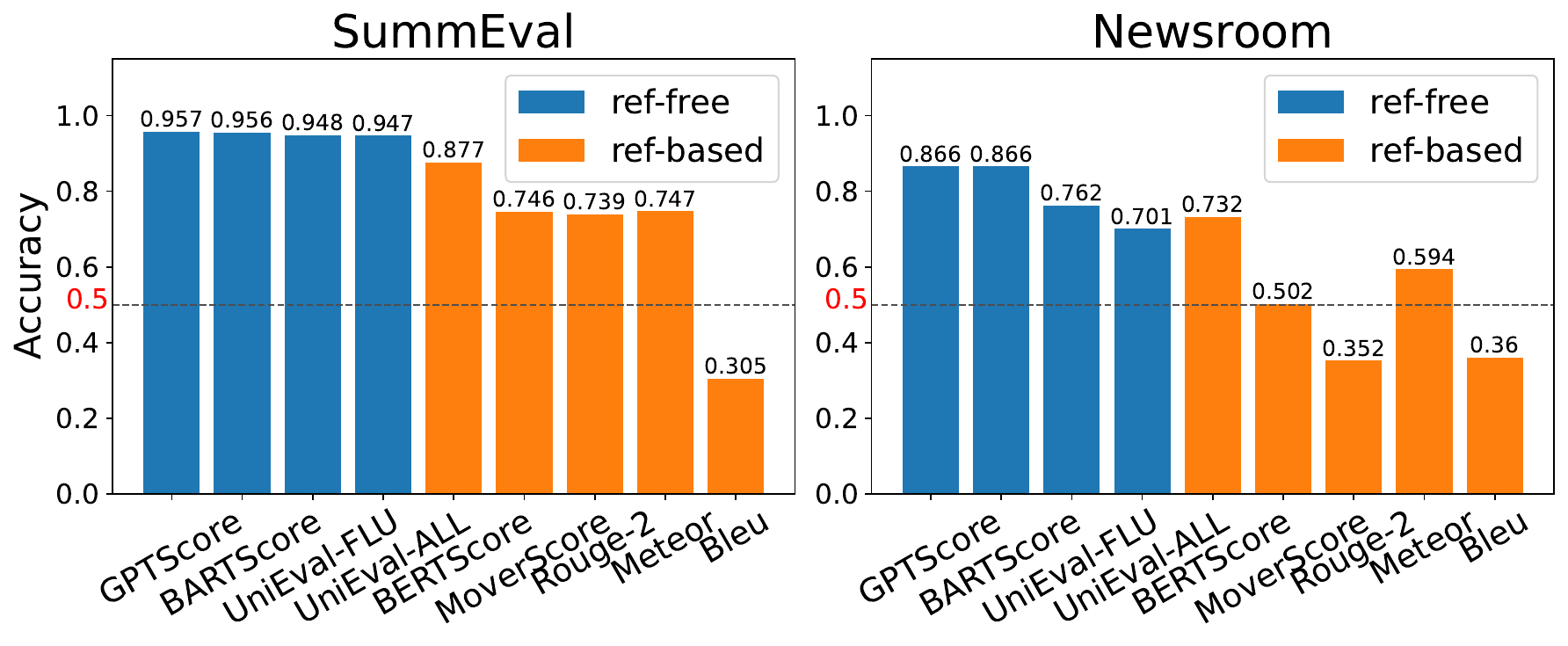} % 插入第二个子图
        \subcaption{Fluency}
        \label{fig:flu_perturb_acc}
    \end{minipage}
    \caption{Accuracy of detecting perturbation with each metric.  Here accuracy is defined as the proportion of text pairs where the original score $s$ satisfies $s > \hat{s}$. }
    \vspace{-0.25cm}
\end{figure}

Following the methodology of~\citet{sai-etal-2021-perturbation}, We employ "sentence reorder" and "subject-verb disagreement" techniques for coherence and fluency perturbation, respectively. We use the proportion of text pairs where the original score $s$ satisfies $s > \hat{s}$ as the statistical index to evaluate the capability of metric $M$ in detecting perturbations. The results of coherence and fluency perturbation are depicted in Figure~\ref{fig:coh_perturb_acc} and Figure~\ref{fig:flu_perturb_acc}, respectively. 

We observe that GPTScore and BARTScore outperform other metrics on both criteria and datasets. The performance of UniEval on fluency is relatively worse but also outperforms other reference-based metrics. In comparison, the outcomes of reference-based metrics on detecting coherence are unsatisfying. On coherence perturbation detection. ROUGE, Meteor, and BLEU could only obtain a score far below 50\%, which is the expected accuracy of random selection. The reason should be that these metrics solely focus on surface-level n-gram features and cannot distinguish changes in shuffling sentences, as they provide the same score for both original and perturbed text. BERTScore and MoverScore exhibit better capability but are also not competitive with reference-free metrics.

We owe the weakness of reference-based metrics to the lack of semantic information contained in the embedding distance or n-gram difference, for a single reference only provides a possible answer to the given input, while more semantic knowledge is contained in the underlying model of reference-free metrics. 

%meta correlation
\begin{figure*}[t]
    \begin{minipage}[t]{0.33\textwidth} 
        \centering
        \includegraphics[width=\linewidth]{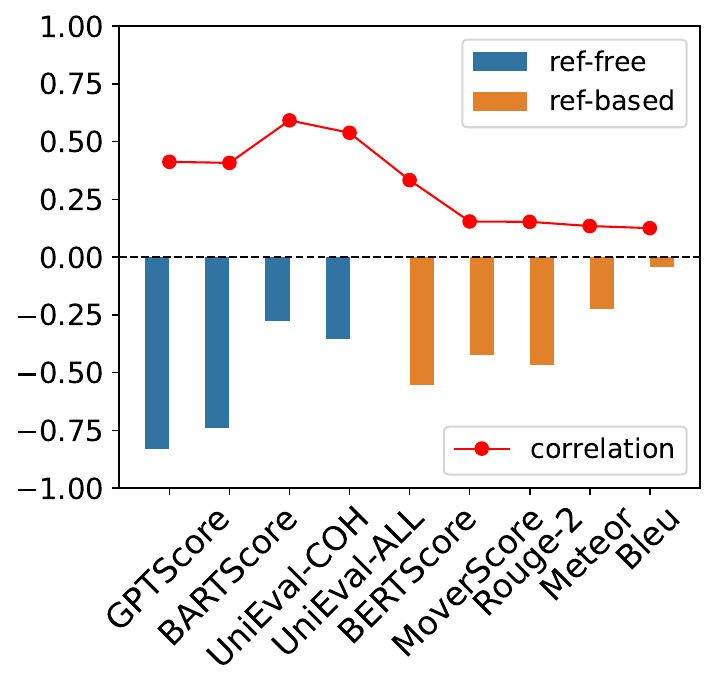}
        \subcaption{Coherence}
        \label{fig:system-coh}
    \end{minipage}%
    \begin{minipage}[t]{0.33\textwidth}
        \centering
        \includegraphics[width=\linewidth]{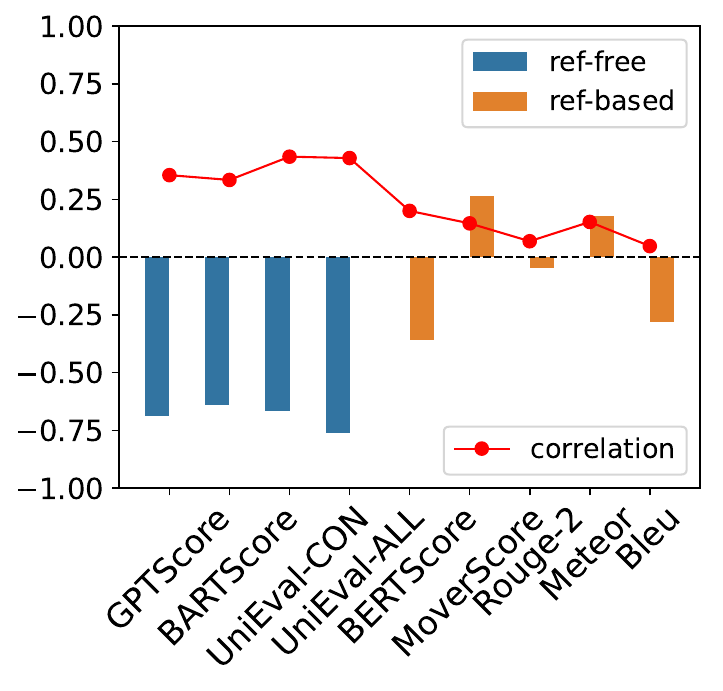} 
        \subcaption{Consistency}
        \label{fig:system-con}
    \end{minipage}
    \begin{minipage}[t]{0.33\textwidth} 
        \centering
        \includegraphics[width=\linewidth]{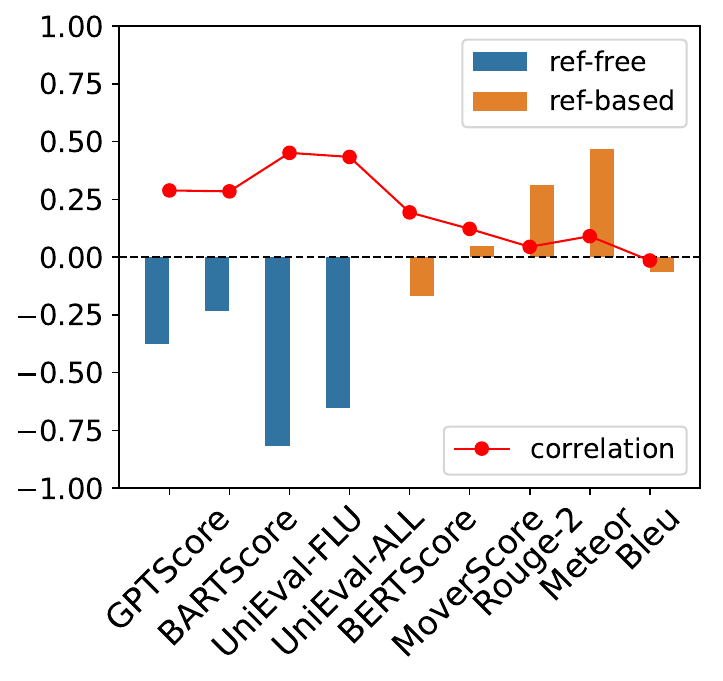} 
        \subcaption{Fluency}
        \label{fig:system-flu}
    \end{minipage}
    \caption{\textbf{Meta-correlation} score for Spearman correlation of each metric on SummEval dataset. The red line represents the Spearman correlation with human judgment obtained in Section~\ref{sec:standard_correlation_test}}
    \label{fig:system_bar}
\end{figure*}

\subsection{Kolmogorov-Smirnov Test}
Based on metrics and criteria mentioned in Table~\ref{tab:dataset_criteria}, we calculate the KS score for each metric on distinguishing sentences with high human-like quality and low human-like quality.
As the score range of human judgment varies across each dataset, the standard of categorizing high-quality and low-quality sentences differs, as outlined in Table~\ref{tab:quality_standard}. 

\begin{table}[h]
    \small
    \centering
    \begin{tabular}{cccc}
    \toprule
    Dataset & Low Quality & High Quality& Range\\
    \midrule
      BAGEL & $<3$ & $\geq$ 5 & $[1,6]$\\
      Newsroom & $<3$ & $\geq$ 4 & $[1,5]$\\
      QAGS & $<1$ & $\geq 1$ & $[0, 1]$\\
      SFHOT & $<3$ & $\geq$ 5 & $[1,6]$\\
      SFRES & $<3$ & $\geq$ 5 & $[1,6]$\\
      SummEval & $<3$ & $\geq$ 4 & $[1,5]$\\
      USR & $<2$ & $\geq$ 2 & $[0,3]$\\
      WebNLG & $<2$ & $\geq$ 3 &$ [1,5]$\\
    \bottomrule
  \end{tabular}
  \caption{Classification standards of high-quality and low-quality 
  sentences for each dataset.}
  \label{tab:quality_standard}
  % \vspace{-0.35cm}
  \end{table}

Results of KS scores are illustrated in Figure~\ref{fig:ks_heatmap} (See raw results in Appendix Table~\ref{tab:ks_all}). In general, reference-free metrics have higher KS scores than reference-based metrics across all three criteria, which indicates a better performance in identifying low-quality texts from high-quality ones. Among reference-based metrics, embedding-based metrics BERTScore and MoverScore have better performance than other n-gram-based metrics. This \textbf{aligns with} the correlation scores presented in Table~\ref{tab:spearman_all}, where metrics with higher correlation scores generally exhibit better capabilities in distinguishing high-quality from low-quality sentences, and vice versa.

\subsection{Stability Analysis}
\label{sec:quality_test}
We further investigate the relationship between system quality and metrics' human correlation score following~\cite{shen2023large}, as introduced in Section~\ref{sec:system_level_analysis}. The outcome of the meta-correlation calculated with Spearman correlation is presented in Figure~\ref{fig:system_bar} (See raw data in Appendix Table~\ref{tab:meta_raw}). We also include the result of correlations with human judgment shown in Table~\ref{tab:spearman_all} in the figure for better comparison. 

We observe that all reference-free metrics and BERTScore have negative meta-correlation scores, and these metrics are also the ones that have the highest correlations with human judgment. In contrast, meta-correlation scores for the rest reference-based metrics differ widely on each criterion. 

This indicates that as the quality of sentences increases, the assessment provided by reference-free metrics has a weaker correlation with human judgment, and their performance is not stable on different criteria. 
Considering the results of the criterion-level analysis in Section~\ref{sec:perturb}, reference-free metrics are capable of identifying lower-quality sentences and assigning lower scores to them, but may not be reliable for handling texts with high quality. Similar descriptions are also suitable for reference-based metrics.

\section{How to better utilize automatic metrics? }
In this section, we discuss how to appropriately apply automatic evaluation metrics based on observed phenomena in experiments.

If researchers want to directly apply automatic metrics to evaluation:
\begin{itemize}
    \item On task summarization and data-to-text, we suggest using reference-free metrics. 
    \item On task dialogue, we suggest Unieval or BERTScore, depending on the availability of human references.
    \item On new tasks, we suggest researchers use metrics independent of source texts, such as UniEval for fluency evaluation. This can reduce the influence caused by new input and new contextual information.
\end{itemize}
 
If it's possible to collect some sample sentences with human judgment, conducting a pre-assessment before applying metrics to a new task is a good choice. Here we provide an example. 

\begin{enumerate}
    \item Researchers can randomly select 50 generation texts, collect human judgments, and use each metric to generate evaluation scores.
    \item Next, for the analysis of metrics' overall performance, calculate the Spearman correlation of human judgment and metrics output. 
    \item If the correlation is over 0.3, the scoring results of the metrics could be considered as moderately correlated with human judgment.
\end{enumerate}
 We select 0.3 as the threshold of correlation score because, as observed in Section~\ref{sec:standard_correlation_test}, most highest correlations on datasets in this study are over 0.3. A more detailed experiment on the selection of sample numbers is introduced in Appendix~\ref{app:sample_num}, where we find that 50 samples can effectively reflect the performance of metrics. More accurate pre-assessment strategies necessitate additional experimental validation and can be set aside for future investigations.

If metrics are not effective enough, a possible solution is to perform task-specific fine-tuning. Regarding criteria indifferent to application scenarios, developing metrics independent of source text as inputs may decrease the influence of tasks. 

It's also worth noting that, although automatic metrics have developed quickly in recent years, having them replace human assessors still has a long way to go. When it comes to fine-grained, high-quality text evaluation tasks, their assessment results should be taken as reference only.

\section{Related Work}

The rapid growth of NLG techniques and the emergence of LLMs have highlighted the importance of automatic evaluation. Numerous metrics have been developed and are widely used in a great variety of tasks. Apart from metrics used in this study, ~\citet{sai-2022-survey} presents a thorough survey of common evaluation metrics for NLG systems. 

Assessing the effectiveness of automatic metrics therefore becomes an important task, and various meta-eval approaches are proposed. Correlation with human judgment is widely applied, however, as it only provides an evaluation of metrics' overall performance, more fine-grained analyses are developed. For example, \citet{nimah-etal-2023-nlg} and \citet{fomicheva-specia-2019-taking} present meta-eval approaches beyond correlation with human judgment. \citet{sai-etal-2021-perturbation} presents a thorough perturbation template for deeper investigating metrics' ability to detect quality defections. OpenMeva~\cite{guan2021openmeva} focuses on story generation, providing a test suite for meta-evaluation from multiple dimensions, pointing out that many metrics have a poor ability to perceive discourse-level incoherence. 

Comprehensive comparison and analysis of automatic metrics are also of importance, which is also the focus of this work. \citet{callison-burch-etal-2006-evaluating} shows that BLEU is not sufficient for the quality evaluation in the translation task. TRUE~\cite{honovich-etal-2022-true-evaluating} focuses on the evaluation of consistency, explicitly defines the meaning, and provides a standard benchmark. 
\citet{deutsch-etal-2022-limitations} select three reference-free metrics for evaluating machine translation and summarization, indicating that reference-free metrics tend to give texts similar to the output of the underlying model higher scores, instead of human-written sentences, and recommending that reference-free metrics should be used as diagnostic tools instead of evaluation metrics. 
Compared with these researchs, we broaden the types of metrics, criteria and application scenarios, verify the pros and cons of each automatic metric by experiments, and provide possible solutions.
\vspace{-0.05cm}

\section{Conclusion}
In this study, we aim to provide insights into the appropriate usage of automatic evaluation metrics. To achieve this goal, we thoroughly examine the performance of reference-based and reference-free metrics with various meta-analysis methods. Our experiments show that, compared with reference-based metrics, the evaluation results provided by reference-free metrics have a closer correlation with human judgment. Also, reference-free metrics are more sensitive to the semantic deficiency in texts. However, the performance of reference-free metrics is task-dependent and is not stable as the quality of candidate texts increases. Therefore, we recommend assessing metrics before applying them to new tasks and new criteria, especially when metrics are not explicitly designed to be used in the specific scenario.

\section*{Limitations}
\begin{itemize}
    \item This work focuses on evaluation experiments conducted on three specific tasks due to limited data availability for specific criteria and human annotations, and the language is restricted to English as well. Further investigation is necessary to validate the discovered performance on more tasks and languages.
    
    \item  The analysis presented in this study is grounded in experiments, however, theoretical analysis is lacking to augment the findings. Regrettably, we did not incorporate mathematical analysis explaining the underlying mechanisms and rationales behind the limitations inherent to each metric. Such mathematical analysis is valuable in the applications of automatic metrics within new domains.
    \item Regarding the weakness of automatic metrics revealed in this study, it's regrettable that corresponding solutions are proposed but are not fully validated. Future work focusing on improving metrics' performance is required. 
\end{itemize}

\bibliography{reference}

\clearpage
\appendix

\section{Implementation Details}
\label{app:implementation}
\subsection{Datasets}
\label{app:dataset}

\paragraph{SummEval}
SummEval provides a collection of summarization results generated by language models~\cite{fabbri-etal-2021-summeval}, which is trained on the CNN/DailyMail datasets~\cite{hermann2015teaching} and the corresponding reference texts. For each generated summary, the dataset also contains score results from both expert annotators and crowd-workers, from four dimensions: coherence, consistency, fluency, and informativeness. 

\paragraph{NEWSROOM}
NEWSROOM collects 60 articles and the corresponding summarization results of 7 models, with human-written summaries as references~\cite{grusky2018newsroom}. Evaluation of coherence, fluency, relevance, and informativeness is available. 

\paragraph{QAGS}
QAGS involves reference texts and annotation results for consistency on the summarization task~\cite{wang-etal-2020-asking}. For each sentence in a generated summary, 3 annotations are collected and the majority vote strategy is used to get a consistency score, and the mean value of all sentences is the final score. 

\paragraph{SFHOT and SFRES}
SFHOT and SFRES provide evaluation results on the data-to-text task, with annotation of naturalness and informativeness~\cite{wen-etal-2015-semantically}. Here informativeness measures the uniform degree of sources and hypotheses, and we use this data for analysis on consistency, as well as naturalness for fluency. 

\paragraph{WebNLG}
WebNLG contains the human evaluation results for the WebNLG Challenge held in 2017, which is a data-to-text task~\cite{shimorina2019webnlg}. The candidate text is evaluated from 3 aspects: fluency, grammar, and semantics. Here fluency measures whether a text is fluent and natural, and we use the fluency score for experiments. 

\paragraph{BAGEL}
BAGEL contains annotations on data-to-text tasks collected from a dialogue system, with human annotation from informativeness and naturalness~\cite{mairesse-etal-2010-bagel}. Here informativeness is compared with the gold standard and is different from our definition. We only use the judgment results for naturalness.

\paragraph{USR}
The USR dataset provides evaluation results on the dialogue task from 5 aspects: fluency, coherence, engagingness, groundedness, and understandability. Following the rephrasing strategy of \cite{zhong-etal-2022-unieval}, we rename the original aspects "maintains context" and "natural" as "coherence" and "fluency".

The resources of all datasets we used are listed as follows.
\begin{itemize}
    \item Newsroom, SummEval, QAGS\_cnn, QAGS\_XSUM, SFHOT, SFRES are downloaded from source provided by~\citet{NEURIPS2021_bartscore}. The related url is \url{https://github.com/neulab/BARTScore}.
    \item WebNLG is downloaded from source provided by~\citet{Scialom2021BEAMetricsAB}. The related url is \url{https://github.com/ThomasScialom/BEAMetrics}. We delete empty reference sentences before applying.
    \item USR\_Topical and USR\_Persona are created by\citet{mehri-eskenazi-2020-usr}. The related URL is \url{https://github.com/shikib/usr}
\end{itemize}

\subsection{ Metrics}
\label{app:metrics}
\begin{itemize}
    \item BARTScore is downloaded from \url{https://github.com/neulab/BARTScore}. We use the faithfulness-based variant based on "facebook/bart-large-cnn"\footnote{\url{https://huggingface.co/facebook/bart-large-cnn}} checkpoint~\cite{lewis2020bart}.
    \item BERTScore is downloaded from \url{https://github.com/Tiiiger/bert_score}. We use the F1 score calculated based on checkpoint "deberta-xlarge-mnli"\footnote{\url{https://huggingface.co/microsoft/deberta-xlarge-mnli}}~\cite{he2021deberta}.
    \item GPTScore is downloaded from \url{https://github.com/jinlanfu/GPTScore} and we use the checkpoint "gpt2-large"\footnote{\url{https://huggingface.co/gpt2-large}}~\cite{radford2019language}.
    \item UniEval is downloaded from \url{https://github.com/maszhongming/UniEval}. We use the "summarization" variant developed based on checkpoint "MingZhong/unieval-sum"\footnote{\url{https://huggingface.co/MingZhong/unieval-sum}}~\cite{zhong-etal-2022-unieval}.
    \item For metric BLEU and Meteor, we use the implementation provided by the python package NLTK~\cite{bird2009natural}. 
    \item For metric ROUGE, we use the implementation provided by the python package rouge\footnote{\url{https://pypi.org/project/rouge/}}. In this study, we present the results of ROUGE-2-f. All implementations adhere to the licenses and terms of each artifact and are in alignment with their intended usage.
\end{itemize}

\section{Study on the influences of $ref$s}
\label{app:ref_influence}
Some datasets used in this study contain multiple references and thus provide a good resource to study whether sentence structure would reflect evaluation results. In the SFHOT and SFRES datasets, we compare the evaluation results of RB metrics using different references. We find that, though humans can identify that the references contain the same meaning, the outcomes of reference-based metrics vary greatly. In one case in the SFHOT dataset (id=3), the system output "Can I double check you do not care if dogs are allowed at the hotel?" receives a high score of 0.93 from BERTScore when using the reference "Can I confirm that you do not care if dogs are allowed at the hotel?", while it receives a generally low score of 0.665 with the reference "You do not care whether they allow dogs?". This variation is even worse when using n-gram-based metrics. In another example in the SFRES dataset (id=16), the form of the references is similar, and we obtain stable output from RB metrics. This phenomenon suggests that the evaluation quality of reference-based metrics is highly dependent on the references and poses a potential risk in applications. 

Another piece of evidence is the evaluation result of the Webnlg datasets. The target of the task is transferring a triple into fluent description texts, thus the variation of answers is small, and the reference texts are similar. From the experiment result shown in Table~\ref{tab:spearman_all}, we can see that the performance of reference-based metrics, including the n-gram-based ones, is better than other tasks, obtaining higher correlation scores with human judgments. In comparison, some reference-free metrics have unsatisfying outcomes, probably because they are incapable of handling the input formulation.
\begin{figure*}
    \centering
    \includegraphics[width=0.8\textwidth]{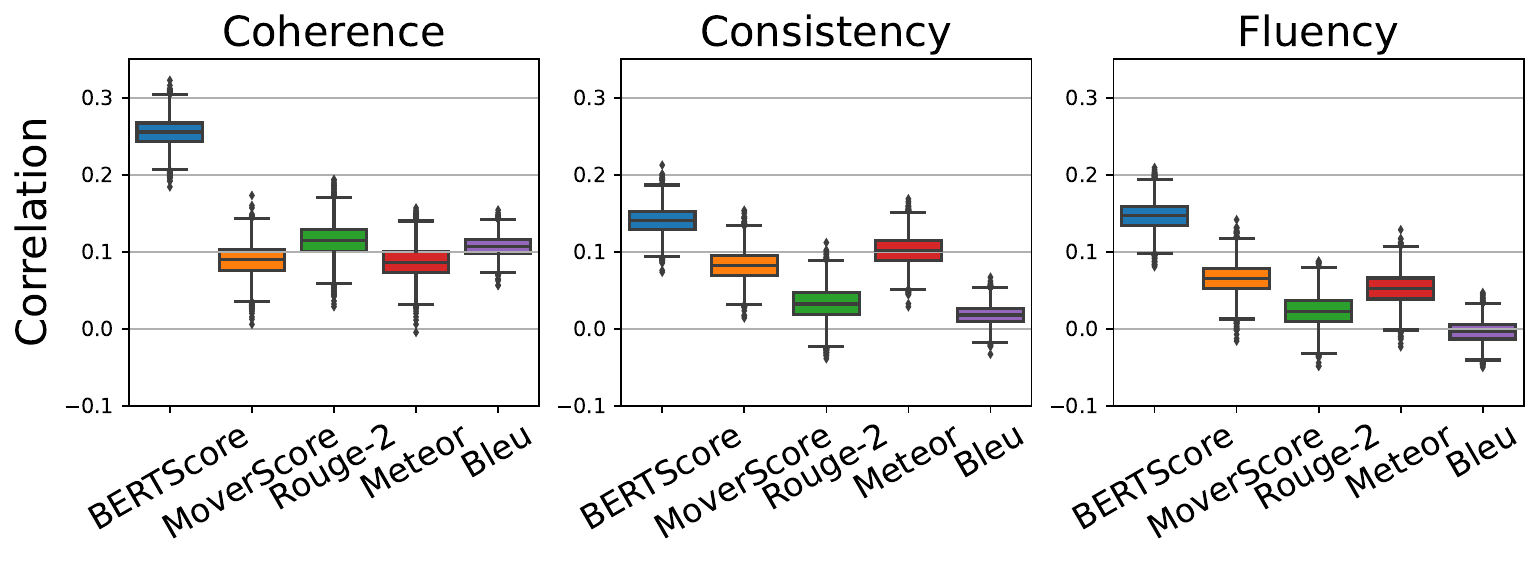}
    \caption{The distribution of correlation scores when using a single reference.}
    \label{fig:single_ref_random_5000}
\end{figure*}

The above examples are all from the data-to-text task, and we also observe similar phenomena in other tasks. Here we use SummEval datasets for further analysis, as it contains 11 references for all 1600 instances, while the reference numbers of other datasets are not so consistent. We randomly select one reference for each instance, calculate the correlation, and repeat this process 5000 times to obtain the distribution of correlation. The results are depicted in Figure~\ref{fig:single_ref_random_5000}.

The wide range of distribution indicates that the selection of references can significantly impact evaluation quality. Therefore, when using reference-based metrics, researchers should be careful in the creation of references, to ensure that the measurement of similarity between $ref$ and $hyp$ can reflect the quality of $hyp$s.

\section{Study of sample number in pre-assessment}
\label{app:sample_num}
This experiment focuses on the number of sample sentences, which is denoted as $n$. On each criterion $c$ and dataset $d$, where $d$ contains evaluation results on $c$, we randomly select $n$ sentences and calculate the Spearman correlation $p_{d, c}$ following Equation~\ref{eq:system_correlation}. We repeat the sampling process for 100 times and calculate the mean value and variance of $p_{d, c}$. The results are shown in Table~\ref{tab:n_sample_20}, Table~\ref{tab:n_sample_50} and Table~\ref{tab:n_sample_100}.

When the sample number is set to 50, the variance of results is relatively small, close to $n=100$, while $n=20$ is not enough. We should also note that the total number of sentences in each dataset is different, while the variance results are similar. A more accurate study on the sampling strategy for pre-assessing is welcomed for future work.

\begin{table*}[]
    
    \small
    \centering
    \tabcolsep 0.05in
    \begin{tabular}{ccccc|ccc|ccc|ccc}
        \toprule
                         &                    & \multicolumn{3}{c|}{\textbf{BARTScore}} & \multicolumn{3}{c|}{\textbf{GPTScore}} & \multicolumn{3}{c|}{\textbf{UniEval-single}} & \multicolumn{3}{c}{\textbf{UniEval-all}} \\
                         \midrule
                         &                    & all      & mean     & std     & all      & mean     & std    & all        & mean      & std       & all       & mean     & std      \\
    \multirow{4}{*}{COH} & Newsroom           & 0.623    & 0.594    & 0.158   & 0.595    & 0.558    & 0.168  & 0.458      & 0.454     & 0.177     & 0.486     & 0.480    & 0.173    \\
                         & SummEval           & 0.408    & 0.390    & 0.196   & 0.412    & 0.410    & 0.203  & 0.592      & 0.557     & 0.158     & 0.538     & 0.522    & 0.167    \\
                         & USR\_persona\ & 0.006    & 0.017    & 0.181   & 0.046    & 0.049    & 0.209  & 0.221      & 0.212     & 0.187     & 0.185     & 0.181    & 0.188    \\
                         & USR\_topic\   & 0.046    & 0.024    & 0.211   & 0.072    & 0.064    & 0.203  & 0.380      & 0.349     & 0.200     & 0.296     & 0.242    & 0.226    \\
                         \midrule
    \multirow{7}{*}{CON} & QAGS\_CNN          & 0.680    & 0.660    & 0.123   & 0.583    & 0.564    & 0.174  & 0.575      & 0.569     & 0.144     & 0.633     & 0.627    & 0.136    \\
                         & QAGS\_XSUM         & 0.159    & 0.171    & 0.214   & 0.081    & 0.089    & 0.221  & 0.369      & 0.379     & 0.203     & 0.344     & 0.360    & 0.211    \\
                         & SFHOT              & 0.222    & 0.233    & 0.205   & 0.219    & 0.216    & 0.208  & 0.185      & 0.185     & 0.213     & 0.270     & 0.293    & 0.198    \\
                         & SFRES              & 0.254    & 0.268    & 0.224   & 0.271    & 0.299    & 0.210  & 0.238      & 0.239     & 0.215     & 0.283     & 0.278    & 0.188    \\
                         & SummEval           & 0.334    & 0.346    & 0.219   & 0.355    & 0.368    & 0.215  & 0.415      & 0.415     & 0.174     & 0.429     & 0.433    & 0.173    \\
                         & USR\_persona\ & -0.019   & -0.005   & 0.201   & -0.099   & -0.080   & 0.226  & 0.063      & 0.030     & 0.231     & 0.050     & 0.014    & 0.246    \\
                         & USR\_topic\   & -0.125   & -0.133   & 0.223   & -0.189   & -0.194   & 0.230  & 0.181      & 0.167     & 0.230     & 0.127     & 0.105    & 0.264    \\
                         \midrule
    \multirow{8}{*}{FLU} & BAGEL              & 0.241    & 0.244    & 0.240   & 0.152    & 0.200    & 0.219  & 0.282      & 0.293     & 0.234     & 0.309     & 0.313    & 0.222    \\
                         & Newsroom           & 0.596    & 0.591    & 0.164   & 0.565    & 0.560    & 0.156  & 0.486      & 0.467     & 0.182     & 0.516     & 0.507    & 0.179    \\
                         & SFHOT              & 0.164    & 0.144    & 0.244   & 0.135    & 0.105    & 0.248  & 0.138      & 0.112     & 0.252     & 0.324     & 0.327    & 0.221    \\
                         & SFRES              & 0.226    & 0.189    & 0.225   & 0.229    & 0.200    & 0.212  & 0.153      & 0.117     & 0.242     & 0.323     & 0.296    & 0.218    \\
                         & SummEval           & 0.285    & 0.300    & 0.233   & 0.288    & 0.302    & 0.236  & 0.346      & 0.331     & 0.227     & 0.434     & 0.409    & 0.235    \\
                         & USR\_persona\ & 0.034    & 0.054    & 0.220   & -0.030   & 0.003    & 0.231  & 0.362      & 0.376     & 0.177     & 0.367     & 0.367    & 0.189    \\
                         & USR\_topic\   & 0.027    & 0.042    & 0.214   & 0.087    & 0.112    & 0.229  & 0.425      & 0.394     & 0.185     & 0.395     & 0.374    & 0.177    \\
                         & webnlg             & 0.330    & 0.330    & 0.227   & 0.072    & 0.058    & 0.252  & 0.480      & 0.475     & 0.180     & 0.565     & 0.555    & 0.184   \\
                         \bottomrule
    \end{tabular}
    \caption{The mean and standard deviation of correlation scores using reference-free metrics with $n=20$.}
    \label{tab:n_sample_20}
\end{table*}

\begin{table*}[]
    \small
    \centering
    \tabcolsep 0.05in
    \begin{tabular}{ccccc|ccc|ccc|ccc}
        \toprule
                         &                    & \multicolumn{3}{c|}{\textbf{BARTScore}} & \multicolumn{3}{c|}{\textbf{GPTScore}} & \multicolumn{3}{c|}{\textbf{UniEval-single}} & \multicolumn{3}{c}{\textbf{UniEval-all}} \\
                         \midrule
                         &                    & all      & mean     & std     & all      & mean     & std    & all        & mean      & std       & all       & mean     & std      \\
    \multirow{4}{*}{COH} & Newsroom           & 0.623    & 0.620    & 0.087   & 0.595    & 0.593    & 0.082  & 0.458      & 0.450     & 0.109     & 0.486     & 0.477    & 0.105    \\
                         & SummEval           & 0.408    & 0.379    & 0.117   & 0.412    & 0.409    & 0.117  & 0.592      & 0.570     & 0.101     & 0.538     & 0.512    & 0.104    \\
                         & USR\_persona & 0.006    & 0.013    & 0.128   & 0.046    & 0.046    & 0.121  & 0.221      & 0.225     & 0.119     & 0.185     & 0.182    & 0.121    \\
                         & USR\_topic   & 0.046    & 0.047    & 0.118   & 0.072    & 0.077    & 0.122  & 0.380      & 0.374     & 0.099     & 0.296     & 0.285    & 0.114    \\
                         \midrule
    \multirow{7}{*}{CON} & QAGS\_CNN          & 0.680    & 0.667    & 0.075   & 0.583    & 0.561    & 0.091  & 0.575      & 0.553     & 0.081     & 0.633     & 0.610    & 0.081    \\
                         & QAGS\_XSUM         & 0.159    & 0.132    & 0.138   & 0.081    & 0.061    & 0.143  & 0.369      & 0.363     & 0.104     & 0.344     & 0.340    & 0.106    \\
                         & SFHOT              & 0.222    & 0.220    & 0.135   & 0.219    & 0.228    & 0.134  & 0.185      & 0.188     & 0.153     & 0.270     & 0.266    & 0.118    \\
                         & SFRES              & 0.254    & 0.251    & 0.125   & 0.271    & 0.259    & 0.120  & 0.238      & 0.243     & 0.123     & 0.283     & 0.272    & 0.118    \\
                         & SummEval           & 0.334    & 0.299    & 0.128   & 0.355    & 0.325    & 0.114  & 0.415      & 0.387     & 0.115     & 0.429     & 0.393    & 0.126    \\
                         & USR\_persona & -0.019   & -0.017   & 0.134   & -0.099   & -0.095   & 0.138  & 0.063      & 0.068     & 0.124     & 0.050     & 0.052    & 0.122    \\
                         & USR\_topic   & -0.125   & -0.139   & 0.123   & -0.189   & -0.186   & 0.138  & 0.181      & 0.166     & 0.122     & 0.127     & 0.115    & 0.126    \\
                         \midrule
    \multirow{8}{*}{FLU} & BAGEL              & 0.241    & 0.255    & 0.137   & 0.152    & 0.162    & 0.133  & 0.282      & 0.301     & 0.129     & 0.309     & 0.316    & 0.129    \\
                         & Newsroom           & 0.596    & 0.596    & 0.104   & 0.565    & 0.569    & 0.107  & 0.486      & 0.499     & 0.110     & 0.516     & 0.529    & 0.098    \\
                         & SFHOT              & 0.164    & 0.151    & 0.133   & 0.135    & 0.122    & 0.138  & 0.138      & 0.157     & 0.147     & 0.324     & 0.335    & 0.131    \\
                         & SFRES              & 0.226    & 0.232    & 0.136   & 0.229    & 0.236    & 0.130  & 0.153      & 0.173     & 0.139     & 0.323     & 0.352    & 0.143    \\
                         & SummEval           & 0.285    & 0.287    & 0.117   & 0.288    & 0.281    & 0.127  & 0.346      & 0.339     & 0.112     & 0.434     & 0.416    & 0.114    \\
                         & USR\_persona & 0.034    & 0.034    & 0.115   & -0.030   & -0.030   & 0.118  & 0.362      & 0.367     & 0.114     & 0.367     & 0.366    & 0.114    \\
                         & USR\_topic   & 0.027    & 0.010    & 0.125   & 0.087    & 0.067    & 0.133  & 0.425      & 0.428     & 0.102     & 0.395     & 0.395    & 0.102    \\
                         & webnlg             & 0.330    & 0.339    & 0.135   & 0.072    & 0.069    & 0.163  & 0.480      & 0.481     & 0.118     & 0.565     & 0.567    & 0.092   \\
                         \bottomrule
    \end{tabular}
    \caption{The mean and standard deviation of correlation scores using reference-free metrics with $n=50$.}
    \label{tab:n_sample_50}
\end{table*}

\begin{table*}[]
    \small
    \centering
    \tabcolsep 0.05in
    \begin{tabular}{ccccc|ccc|ccc|ccc}
        \toprule
                         &                    & \multicolumn{3}{c|}{\textbf{BARTScore}} & \multicolumn{3}{c|}{\textbf{GPTScore}} & \multicolumn{3}{c|}{\textbf{UniEval-single}} & \multicolumn{3}{c}{\textbf{UniEval-all}} \\
                         \midrule
                         &                    & all      & mean     & std     & all      & mean     & std    & all        & mean      & std       & all       & mean     & std      \\
                         \multirow{4}{*}{COH} & Newsroom           & 0.623    & 0.618    & 0.055   & 0.595    & 0.595    & 0.053  & 0.458      & 0.453     & 0.067     & 0.486     & 0.481    & 0.063    \\
                         & SummEval           & 0.408    & 0.413    & 0.077   & 0.412    & 0.413    & 0.085  & 0.592      & 0.584     & 0.063     & 0.538     & 0.534    & 0.071    \\
                         & USR\_persona\ & 0.006    & 0.013    & 0.080   & 0.046    & 0.040    & 0.074  & 0.221      & 0.224     & 0.076     & 0.185     & 0.187    & 0.074    \\
                         & USR\_topic\   & 0.046    & 0.041    & 0.077   & 0.072    & 0.065    & 0.080  & 0.380      & 0.383     & 0.066     & 0.296     & 0.295    & 0.064    \\
                         \midrule
    \multirow{7}{*}{CON} & QAGS\_CNN          & 0.680    & 0.676    & 0.048   & 0.583    & 0.570    & 0.057  & 0.575      & 0.566     & 0.052     & 0.633     & 0.628    & 0.049    \\
                         & QAGS\_XSUM         & 0.159    & 0.154    & 0.083   & 0.081    & 0.078    & 0.080  & 0.369      & 0.362     & 0.071     & 0.344     & 0.337    & 0.074    \\
                         & SFHOT              & 0.222    & 0.212    & 0.091   & 0.219    & 0.211    & 0.090  & 0.185      & 0.170     & 0.103     & 0.270     & 0.268    & 0.092    \\
                         & SFRES              & 0.254    & 0.260    & 0.091   & 0.271    & 0.280    & 0.085  & 0.238      & 0.228     & 0.086     & 0.283     & 0.272    & 0.086    \\
                         & SummEval           & 0.334    & 0.344    & 0.078   & 0.355    & 0.362    & 0.074  & 0.415      & 0.416     & 0.080     & 0.429     & 0.431    & 0.078    \\
                         & USR\_persona\ & -0.019   & -0.021   & 0.078   & -0.099   & -0.104   & 0.084  & 0.063      & 0.077     & 0.078     & 0.050     & 0.058    & 0.077    \\
                         & USR\_topic\   & -0.125   & -0.113   & 0.081   & -0.189   & -0.177   & 0.073  & 0.181      & 0.198     & 0.076     & 0.127     & 0.143    & 0.073    \\
                         \midrule
    \multirow{8}{*}{FLU} & BAGEL              & 0.241    & 0.250    & 0.103   & 0.152    & 0.164    & 0.092  & 0.282      & 0.287     & 0.092     & 0.309     & 0.323    & 0.088    \\
                         & Newsroom           & 0.596    & 0.585    & 0.067   & 0.565    & 0.554    & 0.064  & 0.486      & 0.468     & 0.063     & 0.516     & 0.496    & 0.061    \\
                         & SFHOT              & 0.164    & 0.151    & 0.093   & 0.135    & 0.124    & 0.091  & 0.138      & 0.124     & 0.096     & 0.324     & 0.328    & 0.088    \\
                         & SFRES              & 0.226    & 0.211    & 0.089   & 0.229    & 0.211    & 0.087  & 0.153      & 0.151     & 0.093     & 0.323     & 0.313    & 0.080    \\
                         & SummEval           & 0.285    & 0.278    & 0.098   & 0.288    & 0.290    & 0.101  & 0.346      & 0.336     & 0.093     & 0.434     & 0.426    & 0.092    \\
                         & USR\_persona\ & 0.034    & 0.037    & 0.078   & -0.030   & -0.029   & 0.077  & 0.362      & 0.360     & 0.068     & 0.367     & 0.364    & 0.065    \\
                         & USR\_topic\   & 0.027    & 0.015    & 0.087   & 0.087    & 0.081    & 0.093  & 0.425      & 0.423     & 0.059     & 0.395     & 0.386    & 0.063    \\
                         & webnlg             & 0.330    & 0.329    & 0.091   & 0.072    & 0.075    & 0.101  & 0.480      & 0.477     & 0.075     & 0.565     & 0.560    & 0.074   \\
                         \bottomrule
    \end{tabular}
    \caption{The mean and standard deviation of correlation scores using reference-free metrics with $n=100$.}
    \label{tab:n_sample_100}
\end{table*}

\section{ Supplementary Experiment Results}
\label{app:supplementary}

The raw results of each experiment in this study are listed as follows.
\begin{table*}[]
    \small
    \centering
    \tabcolsep 0.01in
    \begin{tabular}{l@{\hspace{0.1cm}}c@{}c@{\hspace{0.1cm}}c@{\hspace{0.1cm}}c@{\hspace{0.1cm}}c@{\hspace{0.1cm}}|c@{\hspace{0.1cm}}c@{\hspace{0.1cm}}c@{\hspace{0.1cm}}c@{\hspace{0.1cm}}c}
        \toprule
        &           &                \multicolumn{4}{c|}{\textbf{Reference-free}}        & \multicolumn{5}{c}{\textbf{Reference-based}}           \\ 
        \cmidrule{3-11}
  &             &\textbf{ GPTScore} & \textbf{BARTScore} &\textbf{ UniEval} &\textbf{ UniEval\_all} & \textbf{MoverScore }& \textbf{BERTScore} &\textbf{ ROUGE} &\textbf{ Meteor} & \textbf{BLEU}   \\
\midrule
        \multirow{4}{*}{\textbf{COH}}    & Newsroom                    & 0.613                        & \textbf{0.640  }                       & 0.473                       & 0.488                            & 0.070                          & 0.164                         & 0.030                         & 0.108                      & -0.067                   \\
                         & SummEval                    & 0.430                        & 0.434                         & \textbf{0.533}                       & 0.498                            & 0.164                          & 0.349                         & 0.149                         & 0.144                      & 0.011                    \\
                         & USR\_Persona          & 0.040                        & 0.008                         & 0.210                       & 0.173                            & 0.235                          & \textbf{0.259}                         & 0.121                         & 0.188                      & -0.037                   \\
                         & USR\_Topic            & 0.069                        & 0.030                         & 0.385                       & \textbf{0.284}                            & 0.230                          & 0.263                         & 0.188                         & 0.266                      & -0.112                   \\
                         \midrule
    \multirow{5}{*}{\textbf{CON}}    & QAGS\_CNN                   & 0.673                        & \textbf{0.735}                         & 0.635                       & 0.630                            & 0.412                          & 0.585                         & 0.468                         & 0.280                      & 0.071                    \\
                         & QAGS\_XSUM                  & 0.096                        & 0.184                         & \textbf{0.333}                       & 0.317                            & 0.075                          & -0.058                        & 0.121                         & 0.035                      & -0.159                   \\
                         & SFHOT                       & 0.259                        & 0.270                         & \textbf{0.324}                       & 0.324                            & 0.209                          & 0.236                         & 0.101                         & 0.125                      & -0.098                   \\
                         & SFRES                       & 0.316                        & 0.310                         & 0.231                       & \textbf{0.326}                            & 0.196                          & 0.228                         & 0.106                         & 0.193                      & -0.028                   \\
                         & SummEval                    & 0.383                        & 0.377                         &\textbf{ 0.634 }                      & 0.568                            & 0.168                          & 0.227                         & 0.075                         & 0.179                      & 0.026                    \\
                         \midrule
    \multirow{8}{*}{\textbf{FLU}} 
                         & BAGEL                       & 0.268                        & 0.355                         & 0.338                       & \textbf{0.438}                            & 0.205                          & 0.260                         & 0.166                         & 0.121                      & 0.038                    \\
                         & Newsroom                    & 0.571                        & \textbf{0.592}                         & 0.424                       & 0.512                            & 0.050                          & 0.139                         & 0.019                         & 0.079                      & -0.069                   \\
                         & SFHOT                       & 0.167                        & 0.207                         & 0.385                       & \textbf{0.436}                            & 0.177                          & 0.187                         & 0.056                         & 0.043                      & -0.078                   \\
                         & SFRES                       & 0.278                        & 0.282                         & 0.356                       & \textbf{0.394}                            & 0.186                          & 0.216                         & 0.108                         & 0.152                      & 0.042                    \\
                         & SummEval                    & 0.326                        & 0.354                         & 0.597                       & \textbf{0.633}                            & 0.147                          & 0.247                         & 0.055                         & 0.118                      & 0.015                    \\
                         & USR\_Persona          & -0.010                       & 0.039                         & 0.355                       & \textbf{0.398}                            & 0.107                          & 0.326                         & 0.099                         & 0.066                      & -0.120                   \\
                         & USR\_Topic            & 0.106                        & 0.021                         & 0.318                       & \textbf{0.411}                            & 0.201                          & 0.287                         & 0.177                         & 0.227                      & -0.112                   \\
                         & WebNLG                      & 0.093                        & 0.318                         & 0.511                       & \textbf{0.560}                            & 0.424                          & 0.494                         & 0.289                         & 0.338                      & 0.292                    \\
    \bottomrule
    \end{tabular}
    \caption{\textbf{Pearson correlations} of different metrics with human judgments on three criteria in each dataset. Coherence, consistency, and fluency are written in abbreviations COH, CON, and FLU respectively. The \textbf{bold} scores represent the highest correlation results for each task on each criterion.}
    \label{tab:pearson_all}
    \end{table*}

% kendalltau
\begin{table*}[]
    \small
    \tabcolsep 0.01in
    \centering
    \begin{tabular}{l@{\hspace{0.1cm}}c@{}c@{\hspace{0.1cm}}c@{\hspace{0.1cm}}c@{\hspace{0.1cm}}c@{\hspace{0.1cm}}|c@{\hspace{0.1cm}}c@{\hspace{0.1cm}}c@{\hspace{0.1cm}}c@{\hspace{0.1cm}}c}
        \toprule
        &           &                \multicolumn{4}{c|}{\textbf{Reference-free}}        & \multicolumn{5}{c}{\textbf{Reference-based}}           \\ 
        \cmidrule{3-11}
  &             &\textbf{ GPTScore} & \textbf{BARTScore} &\textbf{ UniEval} &\textbf{ UniEval\_all} & \textbf{MoverScore }& \textbf{BERTScore} &\textbf{ ROUGE} &\textbf{ Meteor} & \textbf{BLEU}   \\
\midrule
    \multirow{4}{*}{\textbf{COH}}    & Newsroom                    & 0.440                        & \textbf{0.466  }                       & 0.330                       & 0.351                            & 0.063                          & 0.157                         & 0.061                         & 0.141                      & -0.146                   \\
                         & SummEval                    & 0.297                        & 0.292                         &\textbf{ 0.425   }                    & 0.386                            & 0.109                          & 0.236                         & 0.106                         & 0.096                      & 0.089                    \\
                         & USR\_persona          & 0.035                        & 0.007                         & 0.164                       & 0.138                            & 0.174                          & \textbf{0.195   }                      & 0.083                         & 0.131                      & -0.030                   \\
                         & USR\_topic            & 0.053                        & 0.033                         & \textbf{0.271  }                     & 0.209                            & 0.184                          & 0.216                         & 0.192                         & 0.196                      & -0.126                   \\
    \midrule
                         \multirow{5}{*}{\textbf{CON}}    & QAGS\_CNN                   & 0.470                        & \textbf{0.557}                         & 0.492                       & 0.500                            & 0.278                          & 0.405                         & 0.331                         & 0.256                      & 0.065                    \\
                         & QAGS\_XSUM                  & 0.066                        & 0.130                         & \textbf{0.317}                       & 0.281                            & 0.042                          & -0.047                        & 0.105                         & -0.012                     & -0.136                   \\
                         & SFHOT                       & 0.167                        & 0.170                         & 0.148                       & \textbf{0.206 }                           & 0.154                          & 0.170                         & 0.068                         & 0.053                      & -0.085                   \\
                         & SFRES                       & 0.207                        & 0.193                         & 0.163                       & \textbf{0.215 }                           & 0.130                          & 0.141                         & 0.082                         & 0.134                      & -0.057                   \\
                         & SummEval                    & 0.282                        & 0.264                         & \textbf{0.349}                       & 0.343                            & 0.114                          & 0.157                         & 0.054                         & 0.119                      & 0.038                    \\
    \midrule
                         \multirow{8}{*}{\textbf{FLU}} 
                         & BAGEL                       & 0.110                        & 0.177                         & 0.232                       & \textbf{0.233}                            & 0.139                          & 0.184                         & 0.113                         & 0.079                      & 0.142                    \\
                         & Newsroom                    & 0.419                        & \textbf{0.448}                         & 0.320                       & 0.378                            & 0.031                          & 0.127                         & 0.037                         & 0.110                      & -0.119                   \\
                         & SFHOT                       & 0.098                        & 0.120                         & 0.233                       & \textbf{0.242}                            & 0.114                          & 0.121                         & 0.031                         & 0.010                      & -0.042                   \\
                         & SFRES                       & 0.167                        & 0.165                         & \textbf{0.246 }                      & 0.240                            & 0.113                          & 0.135                         & 0.059                         & 0.106                      & 0.075                    \\
                         & SummEval                    & 0.226                        & 0.223                         & \textbf{0.354}                       & 0.343                            & 0.095                          & 0.151                         & 0.034                         & 0.070                      & -0.012                   \\
                         & USR\_persona          & -0.025                       & 0.026                         & 0.187                       & \textbf{0.289}                            & 0.088                          & 0.248                         & 0.101                         & 0.056                      & -0.096                   \\
                         & USR\_topic            & 0.065                        & 0.018                         & 0.215                       & \textbf{0.284}                            & 0.129                          & 0.208                         & 0.130                         & 0.140                      & -0.065                   \\
                         & WebNLG                      & 0.050                        & 0.238                         & 0.382                       & \textbf{0.415}                            & 0.313                          & 0.367                         & 0.201                         & 0.240                      & 0.232                   \\
    \bottomrule
                        \end{tabular}
                        \caption{\textbf{Kendall's Tau} of different metrics with human judgments on three criteria in each dataset. Coherence, consistency, and fluency are written in abbreviations COH, CON, and FLU respectively. The \textbf{bold} scores represent the highest correlation results for each task on each criterion.}
    \label{tab:kendalltau_all}
    \end{table*}

\begin{table*}[]
    % \small
    \centering
    \begin{tabular}{c|cc|cc}
        \toprule
                     & \multicolumn{2}{c|}{Coherence} & \multicolumn{2}{c}{Fluency} \\
                     \cmidrule{2-5}
                     & Newsroom      & SummEval     & Newsroom     & SummEval    \\
                     \midrule
        GPTScore     & 0.912         & 0.851         & 0.866        & 0.957        \\
        BARTSCore    & 0.932         & 0.851         & 0.866        & 0.956        \\
        Unieval      & 0.667         & 0.791         & 0.762        & 0.948        \\
        Unieval\_all & 0.673         & 0.791         & 0.701        & 0.947        \\
        \midrule
        MoverScore   & 0.585         & 0.628         & 0.502        & 0.746        \\
        BERTScore    & 0.639         & 0.682         & 0.732        & 0.877        \\
        ROUGE        & 0.299         & 0.095         & 0.352        & 0.739        \\
        Meteor       & 0.442         & 0.365         & 0.594        & 0.747        \\
        BLEU         & 0.218         & 0.020         & 0.360        & 0.305        \\
        \bottomrule
    \end{tabular}
    \caption{Accuracy of detecting perturbation with each metric.}
    \end{table*}

\begin{table*}[]
    \small
    \centering
    \tabcolsep 0.02in
    \begin{tabular}{l@{\hspace{0.15cm}}c@{}c@{\hspace{0.1cm}}c@{\hspace{0.1cm}}c@{\hspace{0.1cm}}c@{\hspace{0.1cm}}|c@{\hspace{0.1cm}}c@{\hspace{0.1cm}}c@{\hspace{0.1cm}}c@{\hspace{0.1cm}}c}
      \toprule
      &           &                \multicolumn{4}{c|}{\textbf{Reference-free}}        & \multicolumn{5}{c}{\textbf{Reference-based}}           \\ 
                               \cmidrule{3-11}
                         &             & \textbf{GPTScore} & \textbf{BARTScore} & \textbf{UniEval} & \textbf{UniEval\_all} & \textbf{MoverScore} & \textbf{BERTScore} & \textbf{ROUGE} &\textbf{ Meteor} & \textbf{BLEU}   \\
    \midrule
    \multirow{4}{*}{\textbf{COH}}          & Newsroom    & 0.710    & 0.741     & 0.648       & 0.683       & 0.372     & 0.135      & 0.183   & 0.289  & 0.447 \\
                                  & SummEval    & 0.440    & 0.393     & 0.621       & 0.535       & 0.335     & 0.158      & 0.173   & 0.154  & 0.110 \\
                                  & USR-Persona & 0.058    & 0.122     & 0.271       & 0.232       & 0.238     & 0.212      & 0.066   & 0.105  & 0.044 \\
                                  & USR-Topic   & 0.160    & 0.090     & 0.220       & 0.200       & 0.305     & 0.220      & 0.250   & 0.235  & 0.100 \\
                                  \midrule
    \multirow{5}{*}{\textbf{CON}}          & QAGS\_CNN   & 0.447    & 0.526     & 0.586       & 0.576       & 0.367     & 0.299      & 0.295   & 0.267  & 0.256 \\
                                  & QAGS\_XSUM  & 0.106    & 0.173     & 0.353       & 0.343       & 0.120     & 0.064      & 0.185   & 0.098  & 0.208 \\
                                  & SFHOT       & 0.485    & 0.480     & 0.759       & 0.660       & 0.534     & 0.519      & 0.435   & 0.442  & 0.218 \\
                                  & SFRES       & 0.442    & 0.440     & 0.437       & 0.476       & 0.453     & 0.422      & 0.252   & 0.384  & 0.189 \\
                                  & SummEval    & 0.489    & 0.482     & 0.700       & 0.630       & 0.333     & 0.242      & 0.133   & 0.251  & 0.103 \\
                                  \midrule
    \multirow{8}{*}{\textbf{FLU}}          & BAGEL       & 0.444    & 0.445     & 0.519       & 0.512       & 0.357     & 0.194      & 0.210   & 0.219  & 0.143 \\
                                  & Newsroom    & 0.636    & 0.681     & 0.534       & 0.642       & 0.316     & 0.099      & 0.152   & 0.275  & 0.359 \\
                                  & SFHOT       & 0.348    & 0.389     & 0.559       & 0.577       & 0.318     & 0.296      & 0.150   & 0.128  & 0.161 \\
                                  & SFRES       & 0.371    & 0.358     & 0.555       & 0.564       & 0.326     & 0.283      & 0.200   & 0.289  & 0.218 \\
                                  & SummEval    & 0.435    & 0.485     & 0.784       & 0.733       & 0.402     & 0.223      & 0.128   & 0.171  & 0.209 \\
                                  & USR-Persona & 0.201    & 0.159     & 0.551       & 0.545       & 0.413     & 0.150      & 0.072   & 0.185  & 0.102 \\
                                  & USR-Topic   & 0.231    & 0.149     & 0.261       & 0.294       & 0.229     & 0.186      & 0.149   & 0.204  & 0.083 \\
                                  & WebNLG      & 0.135    & 0.363     & 0.553       & 0.615       & 0.538     & 0.441      & 0.285   & 0.353  & 0.266 \\
    \bottomrule
    \end{tabular}
    
    \caption[]{Kolmogorov-Smirnov (KS) score on distinguishing performance of high-quality and low-quality hyp. }
    \label{tab:ks_all}
\end{table*}

\begin{table*}[]
    \centering
    \begin{tabular}{c|ccc|ccc|ccc}
        \toprule
            \multicolumn{1}{l|}{} & \multicolumn{3}{c|}{Coherence} & \multicolumn{3}{c|}{Consistency} & \multicolumn{3}{c}{Fluency} \\ 
            \cmidrule{2-10}
                                  & Spear.    & Pear.     & Kend.     & Spear.     & Pear.     & Kend.     & Spear.   & Pear.    & Kend.    \\
                                  \midrule
            GPTScore              & -0.829   & -0.787   & -0.683   & -0.688    & -0.871   & -0.483   & -0.378  & -0.412  & -0.243  \\
            BARTScore             & -0.741   & -0.759   & -0.617   & -0.641    & -0.902   & -0.483   & -0.233  & -0.580  & -0.209  \\
            UniEval          & -0.276   & -0.250   & -0.200   & -0.668    & -0.662   & -0.483   & -0.817  & -0.791  & -0.661  \\
            UniEval-ALL           & -0.356   & -0.427   & -0.250   & -0.762    & -0.711   & -0.617   & -0.653  & -0.718  & -0.510  \\
            \midrule
            MoverScore            & -0.424   & -0.455   & -0.367   & 0.265     & -0.012   & 0.200    & 0.047   & -0.436  & 0.092   \\
            BERTScore             & -0.556   & -0.616   & -0.433   & -0.359    & -0.637   & -0.183   & -0.169  & -0.718  & -0.109  \\
            ROUGE-2               & -0.468   & -0.591   & -0.350   & -0.047    & -0.375   & -0.017   & 0.313   & -0.006  & 0.276   \\
            Meteor                & -0.226   & -0.334   & -0.217   & 0.176     & -0.172   & 0.150    & 0.469   & 0.119   & 0.393   \\
            BLEU                  & -0.044   & 0.044    & -0.050   & -0.282    & 0.161    & -0.217   & -0.066  & -0.227  & -0.025 \\

    \bottomrule
    \end{tabular}\caption{Meta-correlation scores of each metric on the SummEval dataset. Spearman correlations, Pearson correlations, and Kendall's Tau are abbreviated as Spear, Pear, and Kend, respectively.}
    \label{tab:meta_raw}
    \end{table*}

\end{document}